\documentclass[letterpaper, 10 pt, conference]{ieeeconf}  

\usepackage{amssymb}
\usepackage{standalone}
\usepackage{algorithm}
\usepackage{algpseudocode}

\usepackage{tikz,lipsum}
\usepackage{lscape}  
\usepackage{booktabs}
\usepackage[utf8]{inputenc}
\usepackage{longtable}
\usepackage[]{color-edits}
\usepackage{makecell}
\usepackage{wrapfig}

\usepackage{cite}

\usepackage{amsmath,amsfonts}
\usepackage{array}
\usepackage{textcomp}
\usepackage{stfloats}
\usepackage{url}
\usepackage{hyperref}
\usepackage{verbatim}
\usepackage{graphicx}

\usepackage{balance}
\usepackage{subcaption}
\usepackage{color, colortbl}
\usepackage{adjustbox}
\usepackage{array}

\IEEEoverridecommandlockouts                              

\title{\LARGE \bf
Semantic Segmentation and Scene Reconstruction of RGB-D Image Frames: An End-to-End Modular Pipeline for Robotic Applications
}

\author{Zhiwu Zheng, Lauren Mentzer, Berk Iskender, Michael Price, Colm Prendergast and Audren Cloitre
\thanks{This work has been included in a filed U.S. patent application (serial number: 63/671,422, filed on 07/15/2024). Z. Zheng and L. Mentzer contributed equally to this work.}
\thanks{The authors are with Analog Garage, Analog Devices, Inc., Boston, MA 02110, USA. 
E-mails: {\tt\scriptsize zhiwu.zheng@analog.com, lauren.mentzer@analog.com, berk.iskender@analog.com, michael.price@analog.com, colm.prendergast@analog.com, audren.cloitre@analog.com}.}
}

\begin{document}

\maketitle
\begin{abstract}
Robots operating in unstructured environments require a comprehensive understanding of their surroundings, necessitating geometric and semantic information from sensor data. Traditional RGB-D processing pipelines focus primarily on geometric reconstruction, limiting their ability to support advanced robotic perception, planning, and interaction. A key challenge is the lack of generalized methods for segmenting RGB-D data into semantically meaningful components while maintaining accurate geometric representations. We introduce a novel end-to-end modular pipeline that integrates state-of-the-art semantic segmentation, human tracking, point-cloud fusion, and scene reconstruction. Our approach improves semantic segmentation accuracy by leveraging the foundational segmentation model SAM2 with a hybrid method that combines its mask generation with a semantic classification model, resulting in sharper masks and high classification accuracy. Compared to SegFormer and OneFormer, our method achieves a similar semantic segmentation accuracy (mIoU of 47.0\% vs 45.9\% in the ADE20K dataset) but provides much more precise object boundaries. Additionally, our human tracking algorithm interacts with the segmentation enabling continuous tracking even when objects leave and re-enter the frame by object re-identification. Our point cloud fusion approach reduces computation time by 1.81× while maintaining a small mean reconstruction error of 25.3 mm by leveraging the semantic information. We validate our approach on benchmark datasets and real-world Kinect RGB-D data, demonstrating improved efficiency, accuracy, and usability. Our structured representation, stored in the Universal Scene Description (USD) format, supports efficient querying, visualization, and robotic simulation, making it practical for real-world deployment.
\end{abstract}


\section{Introduction}
\label{sec:introduction}

Cognitive robots need to understand their environment at the same level of comprehension as humans to be deployed in an unstructured environment. RGB-D data naturally lends itself for robot sensing as it combines what objects are (texture) and where they are (depth) into one data stream. Traditional Simultaneous Localization and Mapping (SLAM) used only geometric information from the data but did not leverage the semantic information \cite{campos2021orb}. Recent approaches have explored integrating semantic information into SLAM, but existing methods face key challenges. Kimera \cite{rosinol2021kimera} uses Mask-RCNN \cite{he2017mask} for semantic labeling, but applies it post hoc, leading to inefficiencies in map generation. iLabel \cite{zhi2022ilabel} and vMAP \cite{kong2023vmap} leverage neural field representations for mapping, but their reliance on learned scene priors can introduce inaccuracies in real-world applications. These methods either suffer from high computational costs, limited generalization, or inaccurate geometric reconstructions.

On the other hand, accurate semantic segmentation of RGB-D data is crucial for enabling robots to understand and navigate complex environments effectively. However, existing methods often face challenges in generalization due to their reliance on training with small datasets that include limited scene types. For example, the SUN RGB-D dataset contains 10,335 RGB-D images in only 37 categories \cite{song2015sun}. The constrained diversity in these datasets can lead to models that perform well in specific scenarios but struggle to generalize across varied environments. In addition, additional features for robotic applications, such as efficient point cloud fusion and human tracking, are lacking in existing methods \cite{jia2024geminifusion, dong2024efficient, yin2023dformer}.

\begin{figure}[t]
    \centering
    \includegraphics[width=\columnwidth]{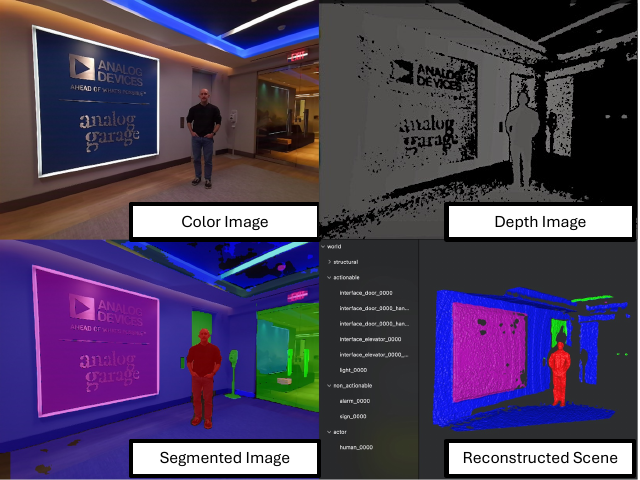}
    \caption{Semantic segmentation and scene reconstruction from color and depth images.}
    \label{fig:segment_and_recon}
\end{figure}

To address these challenges, we introduce a novel pipeline that can segment RGB-D data semantically and transfer this semantic information to reconstruct the scene. Fig. \ref{fig:segment_and_recon} demonstrated an example. The pipeline will take color and depth information (top row) and output its semantic segmented result (bottom left) and finally a 3D reconstructed model of the scene (bottom right). This process leverages semantic information for faster processing of 3D fusion. Furthermore, no assumption is made on the geometry of the reconstructed 3D objects, so that a downstream application can fully trust the reconstructed geometry that comes from measurements without inferences.

The rest of the paper is structured as follows. The pipeline and its components are introduced in Section \ref{sec:pipeline}. Section \ref{sec:results_and_analysis} presents the results of the semantic segmentation, human tracking, scene reconstruction part of the pipeline, and finally an end-to-end test of our entire pipeline with RGB-D data taken from a Kinect sensor. We conclude the paper with insights drawn from our results and possible future work.

\section{Algorithm Pipeline}
\label{sec:pipeline}

\begin{figure*}
    \centering
    \includegraphics[width=\textwidth]{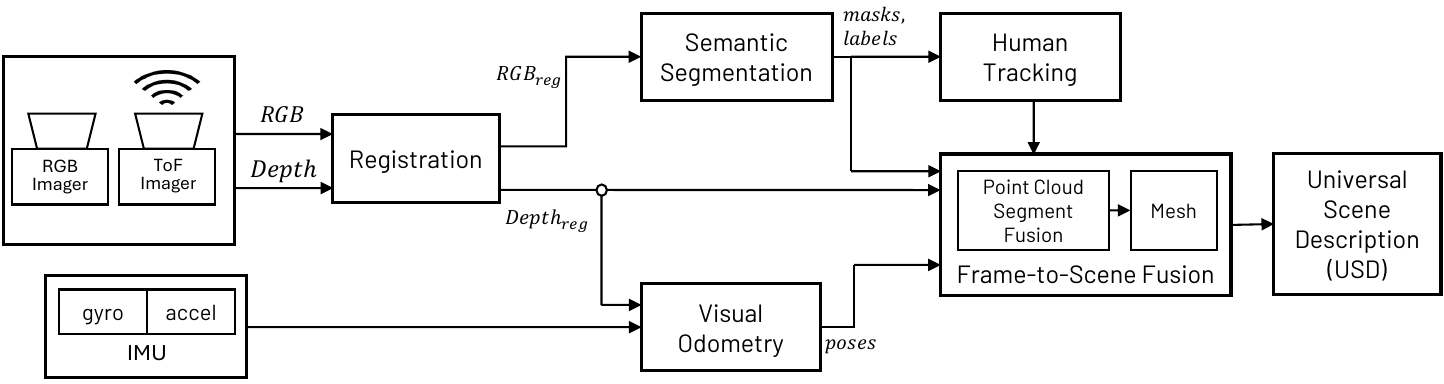}
    \caption{Our pipeline for segmenting and structuring RGB-D data.}
    \label{fig:pipeline}
\end{figure*}

Fig. \ref{fig:pipeline} describes the overall pipeline of our work. RGB-D image frames together with their camera poses were taken with a time-of-flight camera. Semantic segmentation result from the frame was then used for human tracking and point-cloud merging. Human tracking helps point-cloud merging by identifying moving actors. The result is finally stored in the Universal Scene Description (USD) file formation \cite{openusd}. The rest of this section describes these modules in detail.

\subsection{Semantic Segmentation}
\label{sec:segmentation_pipeline}

Segment Anything Model 2 (SAM2) \cite{ravi2024sam} represents the state-of-the-art segmentation algorithm generalized to different types of scenes, but it does not output semantic information. On the other hand, state-of-the-art semantic segmentation algorithms, such as SegFormer \cite{xie2021segformer} and OneFormer \cite{jain2023oneformer}, provide semantic information but less accurate masks. 

\begin{figure}
    \centering
    \includegraphics[width=\columnwidth]{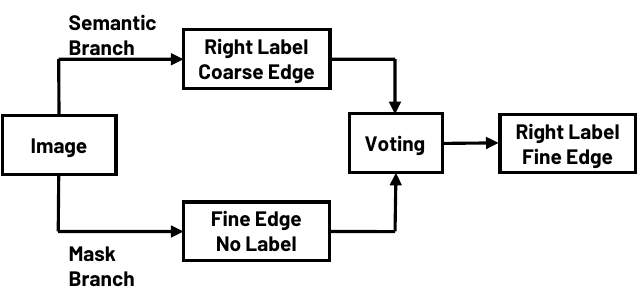}
    \caption{Approach to semantic segmentation. Semantic and mask branches are combined by a voting process that give both right labels and fine edges in masks.}
    \label{fig:segmentation_pipeline}
\end{figure}

Our approach addressed this challenge by leveraging a hybrid architecture \cite{chen2023semantic} with a mask branch implemented by SAM2 and a semantic branch implemented by SegFormer, for both accurate masks and semantic labels. Fig. \ref{fig:segmentation_pipeline} sketches this hybrid approach. The RGB part of a RGB-D image to segment is input to both branches. Pixel-wise classification labels are output from the semantic branch, and segmentation masks are output from the mask branch. A voting process to combine both results is implemented to output new semantic masks and classification labels.

\begin{figure}
    \centering
    \includegraphics[width=\columnwidth]{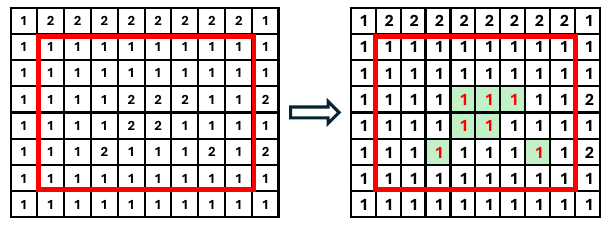}
    \caption{The voting process combines the results of the semantic and mask branches.}
    \label{fig:voting}
\end{figure}

The voting process is described as an example shown in Fig. \ref{fig:voting}, in the following steps: (i) Pixel-wise classification labels were obtained from the semantic branch (e.g. '1' and '2' in Fig. \ref{fig:voting}). (ii) Within each mask from the mask branch, a simple majority vote is used to decide the final classification label (the majority in this example is '1', so all '2's are changed to '1'). (iii) Go to Step (ii) until all the masks are processed. (iv) Finally, for each classification label, output its mask.

\subsection{Human Tracking}
\label{sec:human_tracking}

While SAM2 enables temporal tracking of masks for prompted objects, it does not allow prompting for new objects without resetting its state, losing previously tracked mask IDs. However, continuous tracking of dynamic actors is important for accurate understanding of the surroundings and decision-making. To perform re-identification when SAM2 is reset, we utilize a YOLOX-based \cite{redmon2016you, ge2021yolox} multi-object tracking algorithm with two parts: (i) a bounding box geometry tracking for objects present in consecutive frames, and (ii) SAM2 object pointer-based matching for new detections. We reset SAM2 when the total number of detections changes and assume non-detection of tracked objects and the appearance of new detections do not happen simultaneously. Thus, each event leads to a change in the total number of detections. This is a reasonable assumption for high frame rate scenarios. We note that existing dedicated multi-object tracking algorithms such as \cite{yu2016poi, bewley2016simple, wojke2017simple, zhang2022bytetrack, aharon2022bot, du2023strongsort} can also be incorporated to perform the tracking and re-identification tasks. The focus is on human tracking, but the method can be generalized to different objects.

\textbf{(i) Bounding box geometry tracking.} In line with various methods, including \cite{yu2016poi, bewley2016simple, wojke2017simple, zhang2022bytetrack, aharon2022bot, du2023strongsort}, the Hungarian algorithm \cite{kuhn1955hungarian} is used to match detections across consecutive frames. The elements of the cost matrix $C^{(t)} \in \mathbb{R}^{I \times J}$ in frame $t$, $c_{ij}^{(t)}$, where $i \in \{1,\ldots,I\}$ and $j \in \{1,\ldots,J\}$, represent the cost of assigning the $i$-th detection in the current frame to the $j$-th detection in the previous frame, calculated as the sum of the Euclidean distances between the corner coordinates of the bounding box. This step is performed in every frame to preserve the latest spatial information for the subjects when the SAM2 reset condition is met (that is, $I \neq J$).

\textbf{(ii) SAM2 object pointer tracking.} If the number of detections is larger in the current frame (i.e., $I>J$), unassigned detections after bounding box geometry tracking are compared with a finite memory of SAM2 object pointers from previously tracked subjects using a pre-defined metric, such as the $\ell_2$-norm or cosine similarity. Then, each additional detection is assigned to the distinct instance with the closest object pointer if the proximity (or similarity) exceeds a pre-defined threshold $\tau > 0$. Otherwise, no assignment is performed.
\subsection{Point Cloud Merging}
\label{sec:point_cloud_merging}

As a robot travels through an environment, we seek to enable it to build a representation of the world around it. With each newly captured RGB-D frame, an algorithm is needed for fusing the latest frame into the growing scene representation. This requires associating information in the latest frame with existing scene information. Approaches looking for correspondences at the image level, point cloud level, and mesh level could be considered. Given the availability of accurate pose estimation algorithms and the ease of merging point clouds, we choose an approach of looking for correspondences at the point-cloud level assuming the ability to register data into a common frame. The existing approach \cite{yang2023sam3d} only evaluates the overlap of the registered segmented point clouds for merging. In this work, we incorporate semantic labels to significantly reduce the correspondence search space to the same semantic mask.

More formally, we have the point cloud of the scene in the frame $t-1$ denoted $X^{S_{t-1}}$ which contains $m$ segmented point clouds $\{x_{1}^{s_{t-1}}, \ldots, x_m^{s_{t-1}}\}$ with the corresponding class labels $\{l_{x_{1}^{s_{t-1}}}, \ldots, l_{x_{m}^{s_{t-1}}}\}$. At time $t$ we receive a new RGB-D frame $f_t$ and extract $n$ local segmented point clouds $\{x_1^{f_t}, \ldots, x_n^{f_t}\}$ with corresponding class labels $\{l_{x_1^{f_t}}, \ldots, l_{x_n^{f_t}}\}$. We seek to merge the segmented point clouds in frame $t$ into the segmented point clouds of the existing scene, resulting in an output set of $p$ segmented point clouds $\{x_{1}^{s_t}, \ldots, x_{p}^{s_t}\}$. 

Point clouds representing the same object will overlap, but point clouds representing different objects will have some separation. Leveraging the semantic information not only significantly reduces the search space, but also significantly reduces error in the case that two objects are close to each other.

The overlap evaluation of two point clouds has 3 steps. The given point cloud $x_j^{f_t}$ is first transformed into the same reference frame as the previous merged point clouds in $X^{S_{t-1}}$ using the latest camera pose. Then, a correspondence mapping $M$ is computed between each point in the two point clouds $x_j^{f_t}$ and $x_i^{S_{t-1}}$. Finally, overlap $\sigma_{ij}$ can be estimated as:\begin{equation}
    \label{eq:overlap}
    \sigma_{ij} = \frac{|M|}{\min(|x_i^{S_{t-1}}|, |x_j^{f_t}|)},
\end{equation}
where the $|\cdot|$ operator represents the cardinality of a set.

\begin{figure}
    \centering
    \includegraphics[width=\columnwidth]{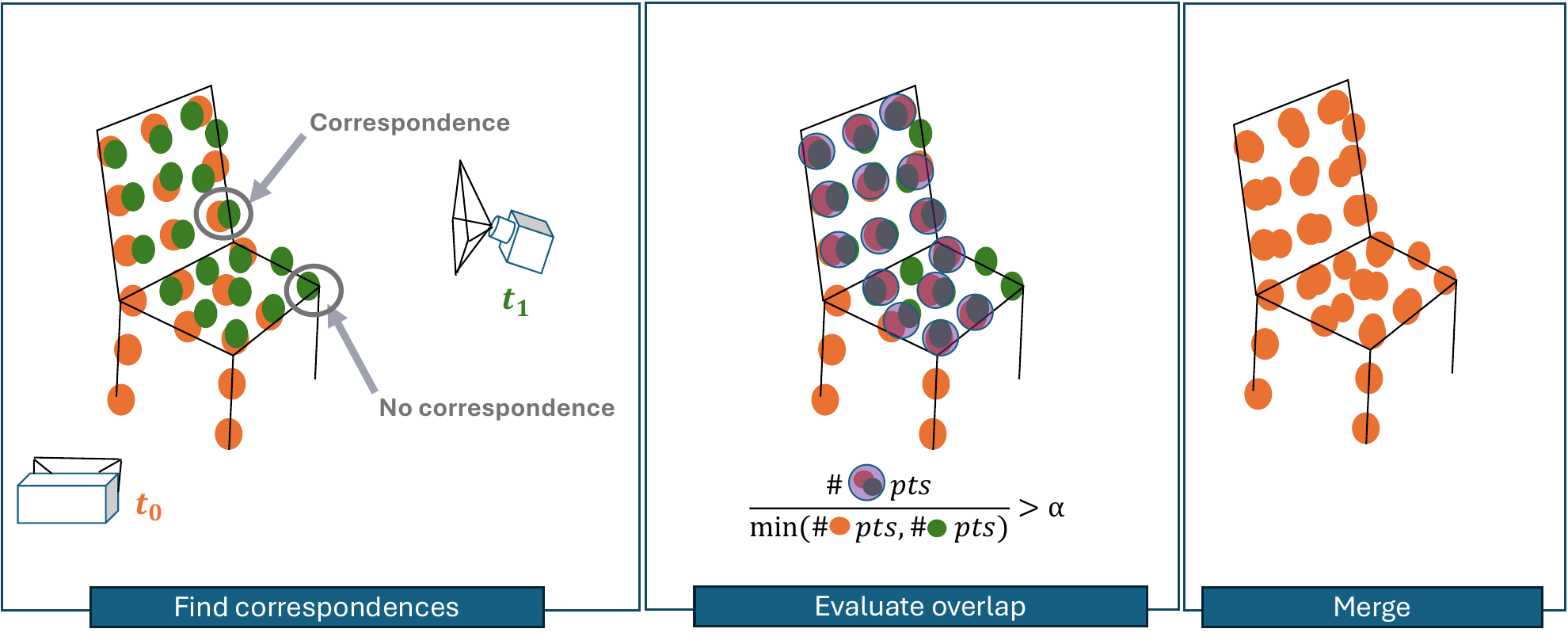}
    \caption{Visualization of the point cloud merging algorithm inspired by SAM3D. Orange and green circles represent 3D points captured at $t=0$ and $t=1$, respectively. Downsampling of the final point cloud is not depicted.}
    \label{fig:pt_cloud_merging}
\end{figure}

Dividing by the smaller of the two point clouds ensures that $\sigma_{ij} \in [0,1]$. If $\sigma_{ij} > \alpha$, $x_i^{S_{t-1}}$ and $x_j^{f_t}$ are declared correspondences and merged into a single point cloud $x_k^{S_t} \in X^{S_t}$. The parameter $\alpha$ can be interpreted as the percentage of points in the smaller point cloud that overlap with the larger point cloud. A visualization of the point-cloud merging algorithm is shown in Fig. \ref{fig:pt_cloud_merging}. We give the following notes on our implementation.

\textbf{(i) Semantics-guided point cloud fusion.} 
We utilize class labels associated with each segmented point cloud. We only evaluate the overlap of $x_i^{S_{t-1}}$ with $x_j^{f_t}$ if $l_{x_i^{S_{t-1}}} = l_{x_j^{f_t}}$. Intuitively, we do not need to consider fusing chair point cloud data with non-chair point cloud data; we only need to check whether we should fuse a chair in $X^{f_t}$ with chairs in $X^{ S_{t-1} }$. These semantic labels could be from a semantic segmentation or an instance segmentation. In Section \ref{sec:results_and_analysis}, we show results using both types of segmentation as input.

\textbf{(ii) Mutually closest points.}
We establish point correspondences as mutually closest points in the two point clouds. We compute the Euclidean distance between each pair of points in $x_i^{S_{t-1}}$ and $x_j^{f_t}$. This results in $|x_i^{S_{t-1}}| |x_j^{f_t}|$ distance computations if done on the raw point clouds. This computation can grow quickly with an increasing point cloud size. To reduce computation time, we do a voxel downsampling before computing point correspondences. This can be understood as placing a voxel grid over the point-cloud data and outputting a point cloud as the centroid of points in each voxel. In our experiments, we use a voxel grid size of 100 millimeters.

\textbf{(iii) Final downsampling.}
After merging point clouds, we do a voxel downsampling on the merged raw point clouds to reduce storing redundant information. In our experiments, this voxel grid size is 50 millimeters. This is smaller than the distance computation voxel grid size, so we can produce a more detailed output point cloud and mesh, while the other is larger, so we can keep just enough information to be able to establish overlap for merging quickly.

After the point clouds are merged, the results are meshed using Open3D's Truncated Signed Distance Function (TSDF) volume integration \cite{Zhou2018}. We re-mesh the updated scene point cloud at each frame and algorithms for only the areas that changed.

Lastly, we represent our scene mesh using the Universal Scene Description (USD) file format invented by Pixar in 2016. USD enables efficient editing of complex scenes by many users, the same file format across simulation and real-time use, utilizing the layered representation it provides, and having a scene representation consistent across many algorithms.

\section{Results and Analysis}
\label{sec:results_and_analysis}

\subsection{Semantic Segmentation}
\label{sec:segmentation_results}
\begin{figure}
\centering
\includegraphics[width=\columnwidth]{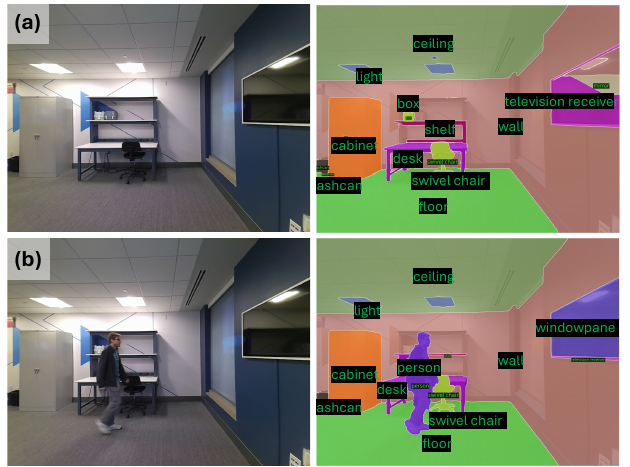}
\caption{Semantic segmentation result of frames (a) and (b) taken from a customized dataset. The left column contains the RGB frames, while the right column shows the corresponding semantic masks and their respective labels.}
\label{fig:segmentation_results}
\end{figure}

Fig. \ref{fig:segmentation_results} illustrates the semantic segmentation results taken customarily as an example. 
SAM2 model \cite{sam2_hiera_large} and the SegFormer trained in ADE20K \cite{segformer_ade} are used as mask and semantic branches, respectively. The left column contains the acquired RGB frames, and the right column overlays the semantic segmentation masks (differentiated by colors) and their semantic labels (texts on top). The resulting masks are accurate and sharp, and the semantic labels are mostly correct. Surprisingly, the desk and the swivel chair that are obscured by the person in the second image are still correctly masked and classified. 

For the rest of this subsection, we compare the differences in semantic accuracy and mask quality with only the semantic branch vs. the combined approach.

\textbf{(i) Semantic accuracy evaluation.}
The accuracy of semantic segmentation is evaluated using the test sets of three datasets, ADE20K \cite{zhou2017scene}, COCO 2017 \cite{lin2014microsoft}, and CityScapes \cite{cordts2016cityscapes}, with SegFormer/OneFormer vs. our hybrid approach. The SAM2 model used in the mask branch was trained on the SA-V and SA-1B datasets\cite{sam2_hiera_large}, and SegFormer/OneFormer is trained on the corresponding training set of the test set\cite{segformer_ade, oneformer_coco_swin_large, segformer_cityscapes}. These datasets cover diversified indoor and outdoor scenes with the scale from a pencil to a mountain. Three metrics are reported for evaluation: mean intersection over union (mIoU), mean class accuracy (mAcc), and pixel accuracy (pAcc).

Table \ref{tab:segmentation_eval} describes the evaluation result. Compared with the semantic branch only, the combined approach does not significantly change the metrics. This might be due to a bottleneck caused by the semantic branch, since the initial classification labels are extracted solely from it.

\begin{table}[h]
    \centering
    \caption{Semantic segmentation accuracies with and without mask branches on ADE20K, COCO2017, and CityScapes datasets.}    \label{tab:segmentation_eval}
    \begin{tabular}{c}
        \includegraphics[width=0.96\linewidth]{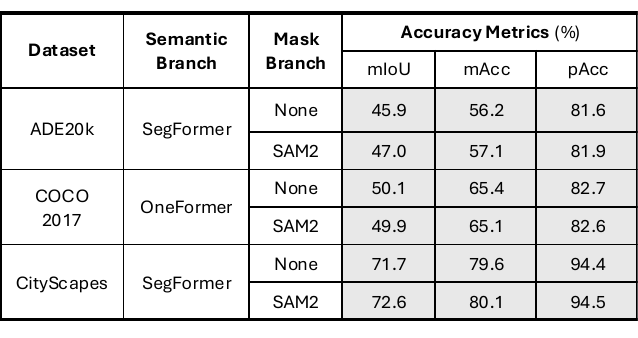}
    \end{tabular}
\end{table}

\textbf{(ii) Mask quality evaluation.}
Although the combined approach does not show clear superiority in semantic accuracy, we observe a significant improvement in mask quality.

\begin{figure}
\centering
\includegraphics[width=\columnwidth]{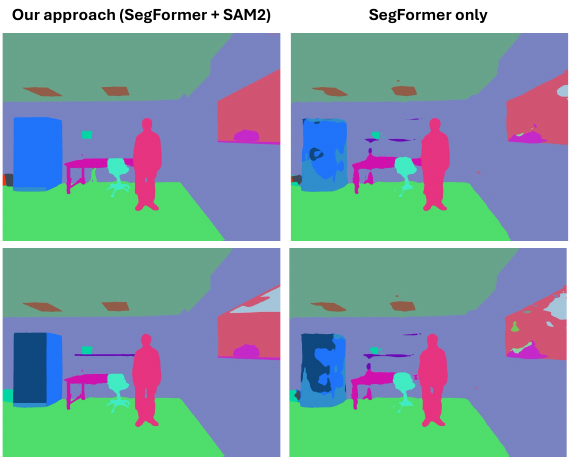}
\caption{Mask outputs for two images (top and bottom) from our approach (left) and using SegFormer only (right).}
\label{fig:mask_quality_comparison}
\end{figure}

Fig. \ref{fig:mask_quality_comparison} illustrates two examples of a scene we captured using our approach vs. SegFormer. The mask outputs in our approach are clearly sharper, more detailed, and more accurate compared with the outputs from only the semantic branch. Quantitative benchmarking of mask qualities is under development.

\subsection{Human Tracking}
\label{sec:human_tracking_results}

In our experiments, we projected 256-dimensional SAM2 object pointers onto a 3-dimensional subspace using PCA and used the Euclidean distance as the similarity metric. The principal components are obtained from the pointers acquired from a distinct frame sequence with multiple subjects. Tests on two different RGB sequences in Fig. \ref{fig:human_tracking_illustration} show that the proposed tracking algorithm preserves accurate spatial information, enabling correct tracking of subjects which leads to accurate assignments of SAM2 masks after the model is reset and re-prompted for new detections. The similarity threshold $\tau$ allows re-identification of previously tracked subjects, tuned by optimizing tracking results using a small set of 7 RGB sequences representing different scenarios.
\begin{table}[hbtp!]
    \centering
    \begin{tabular}{c} 
    \includegraphics[width=\columnwidth]{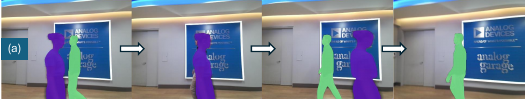} \vspace{-0.05cm} \\
    \hspace{-0.1cm}\includegraphics[width=\columnwidth]{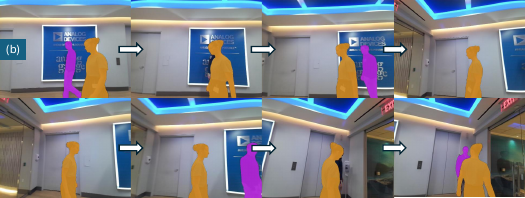}
    \end{tabular}
    \captionof{figure}{Frame results of the tracking algorithm described in Section \ref{sec:human_tracking}. Results are shown for sequences (a) and (b) with 4 and 8 sample frames, respectively. Different colors represent distinct subjects. The horizontal arrows indicate the temporal order.}
    \label{fig:human_tracking_illustration}
\end{table}

\subsection{Point Cloud Merging}
\label{sec:point_cloud_merging_results}

\textbf{(i) Point Cloud Analysis.} 
We evaluate the performance of our point cloud fusion algorithm on the Hypersim dataset \cite{roberts2021hypersim}. Hypersim is a photorealistic synthetic dataset that contains 400+ indoor scenes with RGB, depth, camera poses, and semantic information.

\begin{figure}
    \centering
    \includegraphics[width=\columnwidth]{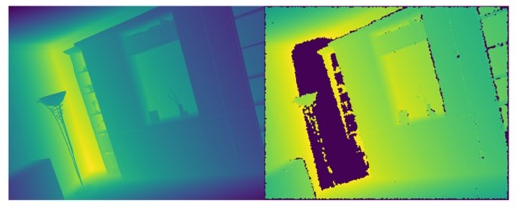}
    \caption{ Ground truth depth provided by Hypersim (left) and the generated noisy depth image to mimic realistic depth sensor data (right). Shown on Hypersim Scene 001-004.}
    \label{fig:pt_cloud_merging_depth_comp}
\end{figure}

A challenge in working with this dataset is the lack of realistic depth images. The provided depth images do not reflect a maximum depth-sensing limitation or noise, unavoidable in real-world data. We add noise to the depth images\cite{handa2014benchmark, barron2013intrinsic, bohg2014robot}, to mimic real hardware noise (e.g., from Kinect). We add a maximum depth filter analogous to real-world depth sensors, set to 4 meters. The result is shown in Fig. \ref{fig:pt_cloud_merging_depth_comp} demonstrating an example of the raw (left) vs. our processed data (right).

\begin{figure*}
    \centering
    \includegraphics[width=\textwidth]{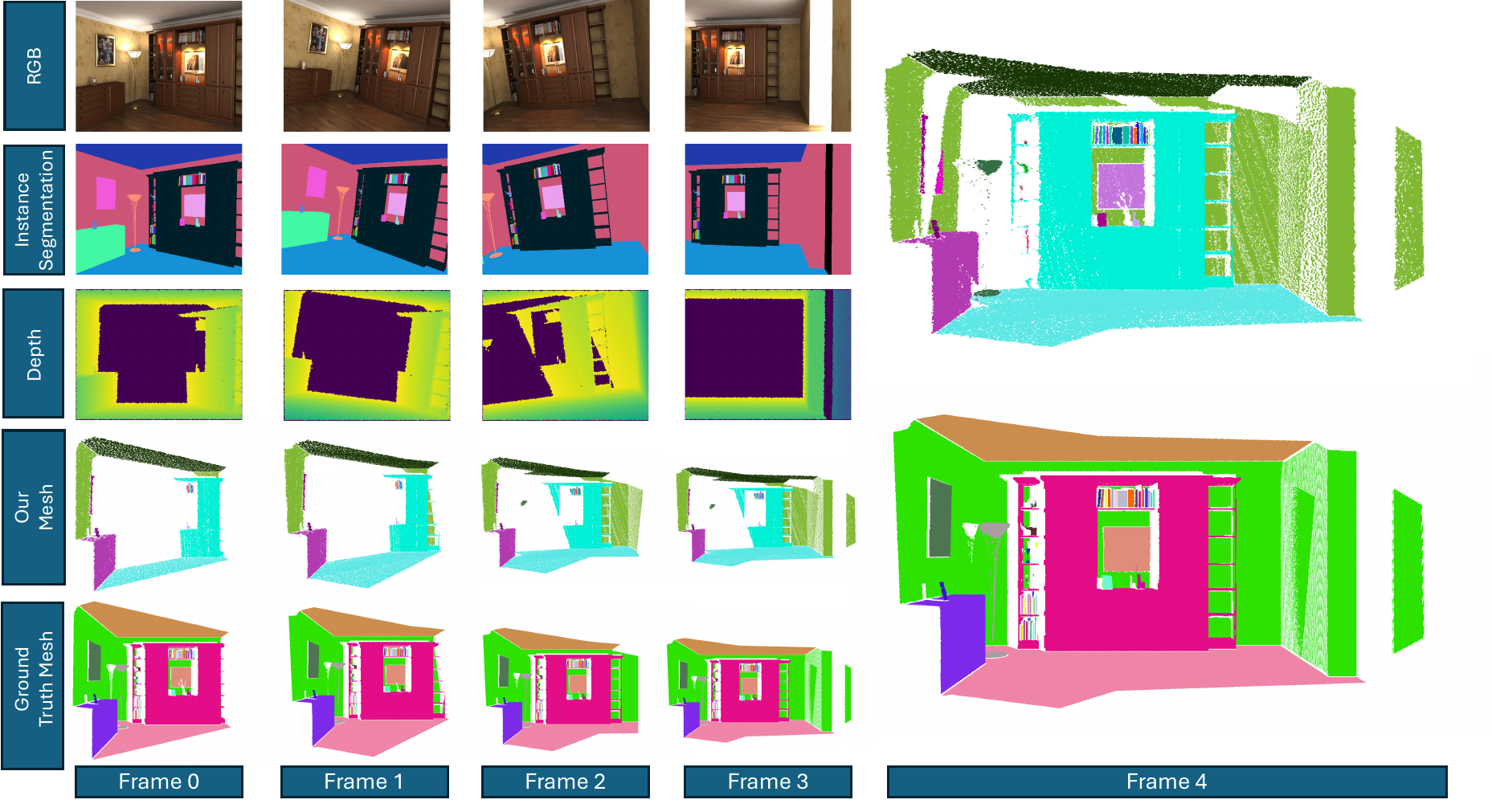}
    \caption{Hypersim scene 001-004 point cloud fusion results.}
    \label{fig:cloud_fusion_res}
\end{figure*}

In Fig. \ref{fig:cloud_fusion_res}, we demonstrated example results of our point cloud merging algorithm on the first few frames in a scene (Scene 001-004) of the Hypersim dataset. The RGB images are provided in Row \#1 for reference. The inputs to the algorithm are instance segmentations (Row \#2) and the processed depth images (Row \#3). The output of our algorithm is shown in Row \#4, where each semantically segmented mesh is shown in a unique color. As new frames are received (indicated by increasing columns from left to right), the point cloud at the latest frame is merged into the merged point cloud of previous frames. From the results of our mesh (Row \#3), we can observe that each object has only one mesh with one unique color and meshes fill in and grow upon re-imaging the same object. For example, in Scene 001-004 Frame \#1, we image the floor from a new viewpoint. Correct point cloud fusion will identify this floor as the same floor from Frame \#0, and we should see one consistent floor mesh in one color as the output of Frame \#1. This means that our fusion is correct since otherwise a single object will have multiple meshes shown in different colors. The result of Frame \# 4 is highlighted showing the mesh from our algorithm (top) vs. the ground truth (bottom). It is very impressive that our mesh can be so close to ground truth after taking just 5 images.

\begin{figure}
    \centering
    \includegraphics[width=\columnwidth]{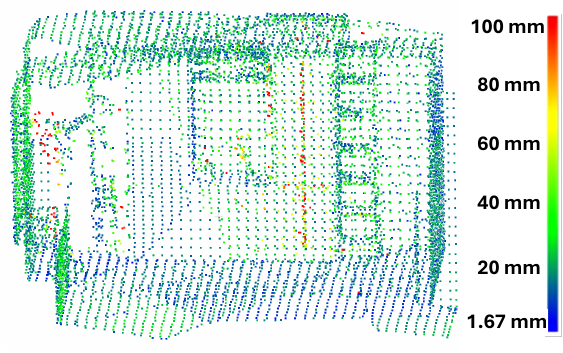}
    \caption{Accuracy analysis of our estimated point cloud compared to the ground truth point cloud.}
    \label{fig:pt_cloud_merging_acc_analysis}
\end{figure}

Finally, Fig. \ref{fig:pt_cloud_merging_acc_analysis} provides a quantitative metric for the accuracy of our fusion for Frame \#4. We use \textit {CloudCompare} \cite{cloudcompare} to calculate the accuracy of our estimated point cloud with respect to the ground truth. The color of the pixel in the estimated point cloud maps to the distance in millimeters to the closest pixel in the ground truth point cloud. The estimated point cloud has a mean error of 25.3 mm and a standard deviation of 19.1 mm. We can see that areas in the scene imaged more times, such as the floor's middle to right side, have lower error, indicated by the blue color.

\textbf{(ii) Runtime Analysis.}
We also evaluate the runtime improvement achieved by incorporating class labels into the point-cloud fusion algorithm, compared to performance without them. In Table \ref{tab:runtime_analysis}, we report the results of an experiment involving merging 6 objects in the indoor scene shown in Fig. \ref{fig:segment_and_recon}, captured with a Kinect sensor. We average the runtimes of point cloud fusion across 10 trials for each frame. Then, we report the average factor reduction in runtime achieved by using labels compared to not using them by computing the runtime ratios. The average improvement across all frames is 1.81$\times$. The benefit of using labels tends to increase with the number of objects with different labels. For example, when merging 2 of the 6 objects in the same scene, we observe an average reduction factor across the 5 frames of only 1.1$\times$. Additionally, the use of labels becomes increasingly advantageous with the number of distance computations that must be performed without them.

\begin{table}[hbtp!]
    \footnotesize
    \setlength{\tabcolsep}{5pt}
    \renewcommand{\arraystretch}{0.5}
    \centering
    \caption{Runtime analysis of point cloud fusion with and without using labels on 6 objects in the scene shown in Fig. \ref{fig:segment_and_recon}.}
    \begin{tabular}{@{}c|ccccc@{}}
    \toprule
    \multicolumn{1}{c}{Frame \#} & \multicolumn{1}{c}{{1}} & \multicolumn{1}{c}{2} & \multicolumn{1}{c}{{3}} & \multicolumn{1}{c}{4} & \multicolumn{1}{c}{{5}} \\
    \midrule
    \makecell{\# of points: scene} & 2850 & 2976 & 3074 & 3100 & 3283 \\
    \midrule
    \makecell{\# of points: frame} & 3842 & 3658 & 3683 & 3597 & 3737 \\
    \midrule
    \makecell{\# of distance computations,\\no labels ($\times10^{6}$)} & 10.95 & 10.87 & 11.32 & 11.15 & 12.27 \\
    \midrule
    \makecell{Average computation reduction\\factor with labels} & 1.83 & 1.52 & 1.97 & 1.59 & 2.15 \\
     \bottomrule
    \end{tabular}
    \label{tab:runtime_analysis}
\end{table}
        
\subsection{End-to-End Test}
\label{sec:kinect_exp}

\begin{figure}[t]
    \centering
    \includegraphics[width=\columnwidth]{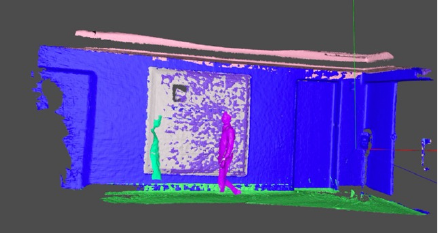}
    \caption{Results of our pipeline in USD on Kinect data. Each color corresponds to a semantic mask for static objects. For humans, each color corresponds to an instance mask, as output by our human tracking pipeline.}
    \label{fig:kinect_results}
\end{figure}

Finally, we evaluate our entire pipeline on data collected from a Kinect RGB-D camera. The video results are contained in our supplementary video submission, and a single frame is shown in Fig. \ref{fig:kinect_results}. Meshes are growing to be more complete with more image input, and each human can be correctly identified thanks to our human tracking algorithm.

\section{Conclusion}
\label{sec:conclusion}

This paper demonstrates a pipeline for segmenting and structuring RGB-D image frames for robotic semantic scene understanding. The pipeline includes semantic segmentation, human tracking, and scene reconstruction with point cloud fusion. 

Our semantic segmentation approach combines semantic and mask branches, implemented with SegFormer/OneFormer and SAM2, respectively. This approach provides accurate masks and correct semantic labels. Moreover, SAM2 does not allow the prompting for new objects after initialization, requiring a reset that results in the loss of previously tracked objects. Our multi-object human tracking algorithm addresses this challenge by combining bounding-box tracking and reassignment using SAM2 object pointers. Furthermore, a point-cloud fusion algorithm is developed by evaluating point overlap per semantic classification to reduce the search space.

Our pipeline is evaluated with standard and self-collected datasets. Compared with SegFormer/OneFormer, our semantic segmentation approach shows statistically on par semantic segmentation accuracy and superior performance in mask qualities. Our human tracking module can track the masks of new and re-appeared objects successfully. Moreover, our point-cloud fusion algorithm shows clearly faster computation speed when utilizing semantic labels and accurate scene reconstruction with a mean error of 25.3 mm.

Future work includes quantitative benchmarking of mask quality and further optimization of similarity matching for tracking.

\bibliographystyle{IEEEtran}
\bibliography{IEEEabrv, refs}

\begin{thebibliography}{10}
\providecommand{\url}[1]{#1}
\csname url@rmstyle\endcsname
\providecommand{\newblock}{\relax}
\providecommand{\bibinfo}[2]{#2}
\providecommand\BIBentrySTDinterwordspacing{\spaceskip=0pt\relax}
\providecommand\BIBentryALTinterwordstretchfactor{4}
\providecommand\BIBentryALTinterwordspacing{\spaceskip=\fontdimen2\font plus
\BIBentryALTinterwordstretchfactor\fontdimen3\font minus \fontdimen4\font\relax}
\providecommand\BIBforeignlanguage[2]{{%
\expandafter\ifx\csname l@#1\endcsname\relax
\typeout{** WARNING: IEEEtran.bst: No hyphenation pattern has been}%
\typeout{** loaded for the language `#1'. Using the pattern for}%
\typeout{** the default language instead.}%
\else
\language=\csname l@#1\endcsname
\fi
#2}}

\bibitem{campos2021orb}
C.~Campos, R.~Elvira, J.~J.~G. Rodr{\'\i}guez, J.~M. Montiel, and J.~D. Tard{\'o}s, ``Orb-slam3: An accurate open-source library for visual, visual--inertial, and multimap slam,'' \emph{IEEE Transactions on Robotics}, vol.~37, no.~6, pp. 1874--1890, 2021.

\bibitem{rosinol2021kimera}
A.~Rosinol, A.~Violette, M.~Abate, N.~Hughes, Y.~Chang, J.~Shi, A.~Gupta, and L.~Carlone, ``Kimera: From slam to spatial perception with 3d dynamic scene graphs,'' \emph{The International Journal of Robotics Research}, vol.~40, no. 12-14, pp. 1510--1546, 2021.

\bibitem{he2017mask}
K.~He, G.~Gkioxari, P.~Doll{\'a}r, and R.~Girshick, ``Mask r-cnn,'' in \emph{Proceedings of the IEEE international conference on computer vision}, 2017, pp. 2961--2969.

\bibitem{zhi2022ilabel}
S.~Zhi, E.~Sucar, A.~Mouton, I.~Haughton, T.~Laidlow, and A.~J. Davison, ``ilabel: Revealing objects in neural fields,'' \emph{IEEE Robotics and Automation Letters}, vol.~8, no.~2, pp. 832--839, 2022.

\bibitem{kong2023vmap}
X.~Kong, S.~Liu, M.~Taher, and A.~J. Davison, ``vmap: Vectorised object mapping for neural field slam,'' in \emph{Proceedings of the IEEE/CVF Conference on Computer Vision and Pattern Recognition}, 2023, pp. 952--961.

\bibitem{song2015sun}
S.~Song, S.~P. Lichtenberg, and J.~Xiao, ``Sun rgb-d: A rgb-d scene understanding benchmark suite,'' in \emph{Proceedings of the IEEE conference on computer vision and pattern recognition}, 2015, pp. 567--576.

\bibitem{jia2024geminifusion}
D.~Jia, J.~Guo, K.~Han, H.~Wu, C.~Zhang, C.~Xu, and X.~Chen, ``Geminifusion: Efficient pixel-wise multimodal fusion for vision transformer,'' \emph{arXiv preprint arXiv:2406.01210}, 2024.

\bibitem{dong2024efficient}
S.~Dong, Y.~Feng, Q.~Yang, Y.~Huang, D.~Liu, and H.~Fan, ``Efficient multimodal semantic segmentation via dual-prompt learning,'' in \emph{2024 IEEE/RSJ International Conference on Intelligent Robots and Systems (IROS)}.\hskip 1em plus 0.5em minus 0.4em\relax IEEE, 2024, pp. 14\,196--14\,203.

\bibitem{yin2023dformer}
B.~Yin, X.~Zhang, Z.~Li, L.~Liu, M.-M. Cheng, and Q.~Hou, ``Dformer: Rethinking rgbd representation learning for semantic segmentation,'' \emph{arXiv preprint arXiv:2309.09668}, 2023.

\bibitem{openusd}
P.~A. Studios, ``{GitHub - PixarAnimationStudios/OpenUSD: Universal Scene Description},'' \url{https://github.com/PixarAnimationStudios/OpenUSD}, 2021.

\bibitem{ravi2024sam}
N.~Ravi, V.~Gabeur, Y.-T. Hu, R.~Hu, C.~Ryali, T.~Ma, H.~Khedr, R.~R{\"a}dle, C.~Rolland, L.~Gustafson, \emph{et~al.}, ``Sam 2: Segment anything in images and videos,'' \emph{arXiv preprint arXiv:2408.00714}, 2024.

\bibitem{xie2021segformer}
E.~Xie, W.~Wang, Z.~Yu, A.~Anandkumar, J.~M. Alvarez, and P.~Luo, ``Segformer: Simple and efficient design for semantic segmentation with transformers,'' \emph{Advances in neural information processing systems}, vol.~34, pp. 12\,077--12\,090, 2021.

\bibitem{jain2023oneformer}
J.~Jain, J.~Li, M.~T. Chiu, A.~Hassani, N.~Orlov, and H.~Shi, ``Oneformer: One transformer to rule universal image segmentation,'' in \emph{Proceedings of the IEEE/CVF Conference on Computer Vision and Pattern Recognition}, 2023, pp. 2989--2998.

\bibitem{chen2023semantic}
J.~Chen, Z.~Yang, and L.~Zhang, ``Semantic segment anything,'' \url{https://github.com/fudan-zvg/Semantic-Segment-Anything}, 2023.

\bibitem{redmon2016you}
J.~Redmon, ``You only look once: Unified, real-time object detection,'' in \emph{Proceedings of the IEEE conference on computer vision and pattern recognition}, 2016.

\bibitem{ge2021yolox}
Z.~Ge, ``Yolox: Exceeding yolo series in 2021,'' \emph{arXiv preprint arXiv:2107.08430}, 2021.

\bibitem{yu2016poi}
F.~Yu, W.~Li, Q.~Li, Y.~Liu, X.~Shi, and J.~Yan, ``Poi: Multiple object tracking with high performance detection and appearance feature,'' in \emph{Computer Vision--ECCV 2016 Workshops: Amsterdam, The Netherlands, October 8-10 and 15-16, 2016, Proceedings, Part II 14}.\hskip 1em plus 0.5em minus 0.4em\relax Springer, 2016, pp. 36--42.

\bibitem{bewley2016simple}
A.~Bewley, Z.~Ge, L.~Ott, F.~Ramos, and B.~Upcroft, ``Simple online and realtime tracking,'' in \emph{2016 IEEE international conference on image processing (ICIP)}.\hskip 1em plus 0.5em minus 0.4em\relax IEEE, 2016, pp. 3464--3468.

\bibitem{wojke2017simple}
N.~Wojke, A.~Bewley, and D.~Paulus, ``Simple online and realtime tracking with a deep association metric,'' in \emph{2017 IEEE international conference on image processing (ICIP)}.\hskip 1em plus 0.5em minus 0.4em\relax IEEE, 2017, pp. 3645--3649.

\bibitem{zhang2022bytetrack}
Y.~Zhang, P.~Sun, Y.~Jiang, D.~Yu, F.~Weng, Z.~Yuan, P.~Luo, W.~Liu, and X.~Wang, ``Bytetrack: Multi-object tracking by associating every detection box,'' in \emph{Proceedings of the European Conference on Computer Vision (ECCV)}, 2022.

\bibitem{aharon2022bot}
N.~Aharon, R.~Orfaig, and B.-Z. Bobrovsky, ``Bot-sort: Robust associations multi-pedestrian tracking,'' \emph{arXiv preprint arXiv:2206.14651}, 2022.

\bibitem{du2023strongsort}
Y.~Du, Z.~Zhao, Y.~Song, Y.~Zhao, F.~Su, T.~Gong, and H.~Meng, ``Strongsort: Make deepsort great again,'' \emph{IEEE Transactions on Multimedia}, 2023.

\bibitem{kuhn1955hungarian}
H.~W. Kuhn, ``The hungarian method for the assignment problem,'' \emph{Naval research logistics quarterly}, vol.~2, no. 1-2, pp. 83--97, 1955.

\bibitem{yang2023sam3d}
Y.~Yang, X.~Wu, T.~He, H.~Zhao, and X.~Liu, ``Sam3d: Segment anything in 3d scenes,'' \emph{arXiv preprint arXiv:2306.03908}, 2023.

\bibitem{Zhou2018}
Q.-Y. Zhou, J.~Park, and V.~Koltun, ``{Open3D}: {A} modern library for {3D} data processing,'' \emph{arXiv:1801.09847}, 2018.

\bibitem{sam2_hiera_large}
``{sam2{\_}hiera{\_}large.pt},'' \url{https://dl.fbaipublicfiles.com/segment_anything_2/072824/sam2_hiera_large.pt}.

\bibitem{segformer_ade}
``{nvidia/segformer-b0-finetuned-ade-512-512},'' \url{https://huggingface.co/nvidia/segformer-b0-finetuned-ade-512-512}.

\bibitem{zhou2017scene}
B.~Zhou, H.~Zhao, X.~Puig, S.~Fidler, A.~Barriuso, and A.~Torralba, ``Scene parsing through ade20k dataset,'' in \emph{Proceedings of the IEEE conference on computer vision and pattern recognition}, 2017, pp. 633--641.

\bibitem{lin2014microsoft}
T.-Y. Lin, M.~Maire, S.~Belongie, J.~Hays, P.~Perona, D.~Ramanan, P.~Doll{\'a}r, and C.~L. Zitnick, ``Microsoft coco: Common objects in context,'' in \emph{Computer Vision--ECCV 2014: 13th European Conference, Zurich, Switzerland, September 6-12, 2014, Proceedings, Part V 13}.\hskip 1em plus 0.5em minus 0.4em\relax Springer, 2014, pp. 740--755.

\bibitem{cordts2016cityscapes}
M.~Cordts, M.~Omran, S.~Ramos, T.~Rehfeld, M.~Enzweiler, R.~Benenson, U.~Franke, S.~Roth, and B.~Schiele, ``The cityscapes dataset for semantic urban scene understanding,'' in \emph{Proceedings of the IEEE conference on computer vision and pattern recognition}, 2016, pp. 3213--3223.

\bibitem{oneformer_coco_swin_large}
``{shi-labs/oneformer{\_}coco{\_}swin{\_}large},'' \url{https://huggingface.co/shi-labs/oneformer_coco_swin_large}.

\bibitem{segformer_cityscapes}
``{nvidia/segformer-b0-finetuned-cityscapes-1024-1024},'' \url{https://huggingface.co/nvidia/segformer-b0-finetuned-cityscapes-1024-1024}.

\bibitem{roberts2021hypersim}
M.~Roberts \emph{et~al.}, ``Hypersim: A photorealistic synthetic dataset for holistic indoor scene understanding,'' in \emph{Proceedings of the IEEE/CVF international conference on computer vision}, 2021, pp. 10\,912--10\,922.

\bibitem{handa2014benchmark}
A.~Handa \emph{et~al.}, ``A benchmark for rgb-d visual odometry, 3d reconstruction and slam,'' in \emph{2014 IEEE international conference on Robotics and automation (ICRA)}.\hskip 1em plus 0.5em minus 0.4em\relax IEEE, 2014, pp. 1524--1531.

\bibitem{barron2013intrinsic}
J.~T. Barron and J.~Malik, ``Intrinsic scene properties from a single rgb-d image,'' in \emph{Proceedings of the IEEE Conference on Computer Vision and Pattern Recognition}, 2013, pp. 17--24.

\bibitem{bohg2014robot}
J.~Bohg \emph{et~al.}, ``Robot arm pose estimation through pixel-wise part classification,'' in \emph{2014 IEEE International Conference on Robotics and Automation (ICRA)}.\hskip 1em plus 0.5em minus 0.4em\relax IEEE, 2014, pp. 3143--3150.

\bibitem{cloudcompare}
``{CloudCompare - 3D point cloud and mesh processing software - open source project},'' \url{https://www.cloudcompare.org}.

\end{thebibliography}

\end{document}